\documentclass{article} 
\usepackage{iclr2016_conference,times}
\usepackage{hyperref}
\usepackage{url}
\usepackage{times}
\usepackage{epsfig}
\usepackage{graphicx}
\usepackage{amsmath}
\usepackage{amssymb}
\usepackage{natbib}
\usepackage{epstopdf}
\usepackage{multirow}

\title{Symmetry-invariant optimization in\\ deep networks}

\author{Vijay Badrinarayanan\\
Department of Engineering\\
Cambridge University\\
Trumpington Street\\
Cambridge, CB2 1PZ, UK\\
\texttt{vb292@cam.ac.uk} \\
\AND
Bamdev Mishra\thanks{This work was initiated while the author was with the Department of Electrical Engineering and Computer Science, University of Li\`ege, 4000 Li\`ege, Belgium and was visiting the Department of Engineering (Control Group), University of Cambridge, Cambridge, UK.} \\
Amazon Development Centre India \\
Bangalore, Karnataka 560055, India \\
\texttt{bamdevm@amazon.com} \\
\AND
Roberto Cipolla \\
Department of Engineering\\
Cambridge University\\
Trumpington Street\\
Cambridge, CB2 1PZ, UK\\
\texttt{roberto@cam.ac.uk} \\
}

\usepackage{arydshln}
\newcommand{\changeBM}[1]{#1} 
\newcommand{\changeVB}[1]{#1} 

\newcommand{\Diag}{{\rm Diag}}
\newcommand{\diag}{{\rm diag}}
\newcommand{\Orth}{{\rm Orth}}
\newcommand{\Trace}{{\rm Tr}}

\begin{document}

\maketitle

\begin{abstract}
 \changeBM{Recent works} have highlighted scale invariance or symmetry that is present in the weight space of a typical deep network and the adverse effect that it has on the Euclidean gradient based stochastic gradient descent optimization. In this work, we show that these and other commonly used deep networks, such as those which use a max-pooling and sub-sampling layer, possess more complex forms of symmetry arising from scaling based reparameterization of the network weights. We then propose two symmetry-invariant gradient based weight updates for stochastic gradient descent based learning. Our empirical evidence based on the MNIST dataset shows that these updates improve the test performance without sacrificing the \changeBM{computational} efficiency of the weight updates. We also show the results of training with one of the proposed weight updates on an image segmentation problem. 
 
\end{abstract}

\section{Introduction}
Stochastic gradient descent (SGD) has been the workhorse for optimization of deep networks \citep{Bottou}. The most well-known form uses Euclidean gradients with a varying learning rate to optimize the weights of a deep network. In this regard, \changeBM{the} recent work \citep{PathSGD} has brought to light simple scale invariance properties or symmetries in \changeBM{the} weight space, which commonly used deep networks possess. These symmetries or \changeBM{invariance to} reparameterizations of the weights \changeBM{imply} that although the \changeBM{loss function} remains \changeBM{invariant}, the Euclidean gradient varies based on the chosen parameterization. \changeBM{In particular, the Euclidean gradient scales inversely to the scaling of the variable \citep{PathSGD}.} This leads to very different trajectories for \changeBM{different reparameterizations of the weights during the training process} \citep{PathSGD}. 


Although these issues have been raised recently, the precursor to these methods is the early work of \citet{Amari}, who proposed the use of \textit{natural gradients} to tackle weight space symmetries in neural networks. The idea is to compute the steepest descent direction for weight update on the manifold defined by these symmetries and use this direction to update the weights. The Euclidean gradient direction which ignores these symmetries is \changeBM{no longer} the steepest descent direction. \changeBM{Recently,} \citet{PascanuNaturalGradient} \changeBM{proposed} a second order method using natural gradients for deep networks. Natural neural networks, on the other hand, define a reparameterization of the network weights such that \changeBM{the} Euclidean and natural gradient based updates are the same \citep{DesjardinsNaturalNN}. They use a block-diagonal approximation of the Fisher information matrix as an approximate \emph{natural metric} \changeBM{(a particular inner product)} to \changeBM{motivate} their \changeBM{proposed} reparameterization \citep{PascanuNaturalGradient}. The works of \citet{OllivierRiemmanianMetricsI, OllivierRiemmanianMetricsII} define several metrics that are also based on approximations of the Fisher information matrix or the Hessian of the loss function to perform scale-invariant optimization. \changeVB{Most of the above-mentioned proposals are either computationally expensive to implement or they need modifications to the architecture.} \changeBM{On the other hand, optimization over a manifold with symmetries has been a topic of much research and provides guidance to other simpler metric choices as we show in this paper \citep{absil08a, mishra14b, boumal15a, journee10a, absil04b, edelman98a, manton02a}.}

\begin{figure*}[t!]
\centering
\includegraphics[width=0.7\textwidth]{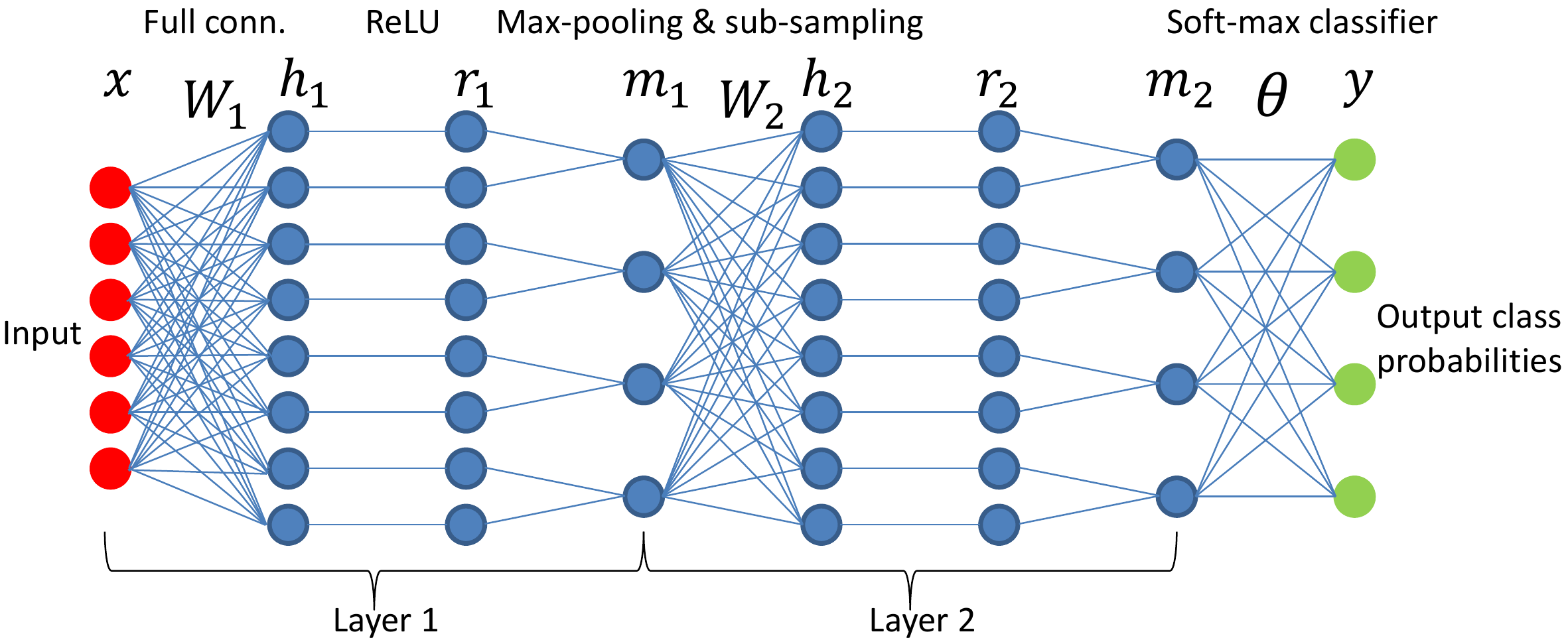}
\caption{Arch1: Deep architecture 1 for classification used in our analysis.}
\label{Arch1}
\end{figure*}
\begin{figure*}[t!]
        \centering
\includegraphics[width=0.7\textwidth]{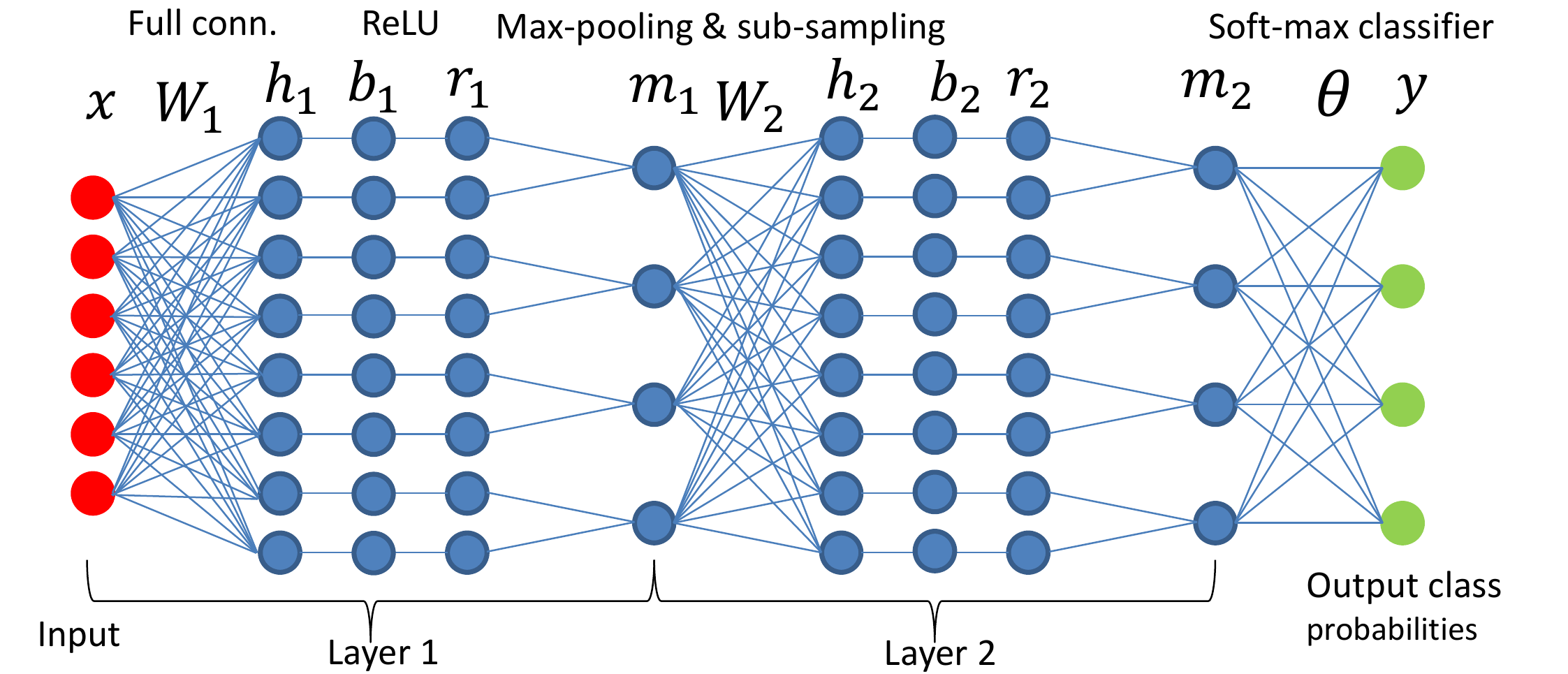}
\caption{Arch2: Deep architecture 2 for classification used in our analysis.}
\label{Arch2}
\end{figure*}

\changeBM{In this paper,} our analysis into some commonly used networks \changeBM{shows that} there \changeBM{exists} more complex forms of symmetries which can affect optimization, and hence there is a need to define \changeBM{\emph{simpler}} weight updates which take into account these invariances. \changeBM{Accordingly, we look at two ways of resolving the symmetries. Both result from a geometric viewpoint on the manifold of the search space. The proposed symmetry-invariant updates are numerically efficient to implement. Even though the focus of the paper is on SGD algorithms, it should be noted that the updates proposed in Table \ref{ProposedUpdates} can readily be extended to first and second order batch algorithms \citep{absil08a}. This paper builds upon and extend our recent work in \citep{badrinarayanan15a}.}

\changeBM{In Section \ref{Architecture},} we analyze the weight space symmetries which exist in a deep architecture commonly for classification, where each layer is composed of a fully connected network, followed by \emph{reLU non-linearity} and a \emph{max-pooling-sub-sampling} step. We then analyze an extension of this architecture, where the output of the fully connected network is batch normalized which has been shown to significantly speed up optimization \citep{BN}. Such architectures in their convolutional form are currently been used for practical problems such as image segmentation, e.g., in \citep{SegNetarXiv}. \changeBM{Section \ref{Optimization} discusses manifold optimization techniques to address the symmetries and we propose simple weight updates and give a motivation behind those. \changeBM{The proposed updates are shown in Table \ref{ProposedUpdates}}. Finally, numerical experiments are discussed in Sections \ref{Experiments}, \ref{Results}, and \ref{DeepConvNets}.}

\changeBM{The stochastic gradient descent algorithms with the proposed updates are implemented in Matlab and Manopt \citep{ManOpt}. The codes are available at \url{http://bamdevmishra.com/codes/deepnetworks}.}


\section{Architectures and symmetry analysis}
\label{Architecture}


To keep the exposition simple, we consider a two layer deep architecture that is shown in Figure \ref{Arch1}. Each layer in Arch1 has typical components commonly found in convolutional neural networks \citep{LeCunNature} such as multiplication with a trainable weight matrix (e.g., $W_{1}$ and $W_{2}$), element-wise rectification ReLU, $2\times1$ max-pooling with stride 2, and sub-sampling. The final layer is a trainable soft-max classifier $\theta$ which predicts the probabilities of the relevant $K$ classes. Arch2 has an additional batch normalization layer when compared to Arch1 \citep{BN}. \changeVB{The analysis in this section readily extends to deeper architectures which use the same components as Arch1 and Arch2.}

\changeVB{The rows of the weight matrices $W_{1}$ and $W_{2}$ correspond to filters in layers $1$ and $2$, respectively. The dimension of each row corresponds to the input dimension of the layer. For example, for the MNIST digits dataset, the input is a $784$ dimensional vector and with $64$ filters in each of the layers, the dimensionality of $W_{1}$ is $64\times784$ and that of $W_{2}$ is $32\times64$. The dimension of $\theta$ is $10\times32$, where each row corresponds to a trainable class vector.}

The element-wise ReLU operation is defined for Arch1 as
\begin{align*}
r_{1,j} = \max(h_{1,j}), \forall j \in h,
\end{align*}
and similarly for Arch2. However, the ReLU operation in Arch2 is performed after the batch normalization (BN) step. The max-pooling and sub-sampling steps are performed as follows,
\begin{align*}
m_{1,k} =  \max(r_{1,k},r_{1,k+1}), \forall k \in m.
\end{align*} 
The cross-entropy objective or loss function for a training example of class $k$ is defined below;
\begin{align*}
L(W_{1},W_{2},\theta) = -\log y_{k}(W_{1},W_{2},\theta).
\end{align*}
During training the loss is summed over a mini-batch of training examples.

Consider the following reparameterization of the trainable parameters
\begin{align}
\label{weightsreparam}
\tilde{W}_{1} =  \alpha_{0} W_{1} \quad {\rm and}\quad  \tilde{W}_{2} =  \beta W_{2}  
\end{align}
for the network in Figure \ref{Arch1} with $8$ filters per layer, where $\alpha_{0}$ is a positive scalar and $\mathbf{\beta} = \Diag (\beta_{1},\beta_{1},\beta_{2},\beta_{2},\beta_{3},\beta_{3},\beta_{4},\beta_{4})$,
where $\beta_{i} > 0$ for $i = \{1,2, 3,4 \}$ and $\Diag(\cdot)$ is an operator which creates a diagonal matrix with its argument placed along the diagonal. \changeBM{It should be noted that there are repeating elements along the diagonal of $\beta$, which comes up because of the max-pooling operation.} \changeBM{Under these} reparameterizations, the changes to the other intermediate outputs in layer $1$ are $\tilde{h_{1}} = \alpha_{0} h_{1}$, $\tilde{r_{1}} = \alpha_{0} r_{1}$, and $\tilde{m_{1}} = \alpha_{0} m_{1}$. \changeBM{Subsequently,} the effect on layer 2 are
$\tilde{h_{2}} = \alpha_{0} \beta h_{2}$, $\tilde{r_{2}} = \alpha_{0} \beta r_{1}$, and $\tilde{m_{2}} = \alpha_{0} \beta_{s} m_{2}$, where $\beta_{s} = \Diag(\beta_{1},\beta_{2},\beta_{3},\beta_{4})$. 

Now, let us reparameterize the class vectors (rows) of the $\theta$ matrix as
\changeVB{
\begin{align}
\label{thetareparam}
\tilde{\theta}_{k} =  \theta_{k} \frac{1}{\alpha_{0}} \beta_{s}^{-1} \quad {\rm for\ } \quad k = \{1,\ldots,K\}, 
\end{align}
}
then evaluating the predicted class probabilities we have,


\[
\displaystyle \tilde{y}_{k} = \frac{ e^{ \tilde{\theta}_{k} \tilde{m}_{2}}} {\sum_{l=1:K}e^{ \tilde{\theta}_{l} \tilde{m}_{2}}}  =  
\changeVB{
\frac{e^{\theta_{k}\frac{1}{\alpha_{0}}\beta_{s}^{-1}\alpha_{0}\beta_{s} m_{2} }}
         {\sum_{l=1:K} e^{\theta_{l} \frac{1}{\alpha_{0}} \beta_{s}^{-1} \alpha_{0} \beta_{s} m_{2} }}  = y_{k}.
         }
\]

Hence, we see that if the weights and classifier parameters are reparameterized as shown in (\ref{weightsreparam}) and (\ref{thetareparam}), then it leaves the loss unchanged. Therefore, there exists continuous symmetries or reparameterizations of $W_1$, $W_2$, and $\theta$ which leave the loss function unchanged. \changeBM{It should be noted that our analysis differs from \citep{PathSGD}, where the authors deal with a simpler case wherein $\beta = \frac{1}{\alpha_{0}}$ \changeBM{is a scalar} and $\theta$ is reparameterization free. It should be emphasized is that the reparameterizations (\ref{weightsreparam}) of $W_1$ and $W_2$ act from the left side (i.e., on the rows) and the reparameterization (\ref{thetareparam}) on $\theta$ acts from the right side (i.e., on the columns).}

The difference between Arch1 and Arch2 is in the introduction of a batch normalization layer in Arch 2 \citep{BN}. Figure \ref{Arch2} shows the network. The idea behind this layer is to reduce the change in distribution of the input features at different layers over the course of optimization so as to speed up convergence. This is accomplished by normalizing each feature (element) in the $h_{1}$ and $h_{2}$ layers to have zero-mean unit variance over each mini-batch. Then a separate and trainable scale and shift is applied to the resulting features to obtain $b_{1}$ and $b_{2}$, respectively. This effectively models the distribution of the features in $h_{1}$ and $h_{2}$ as Gaussians whose mean and variance are learnt during training. Empirical results in \citep{BN} show that this normalization significantly improves convergence and our experiments also support this result. The zero-mean unit-variance normalization of the elements of $h_{1}$ and $h_{2}$ allows for more complex symmetries to exist in the network. Consider the following reparameterizations
\begin{align}
\label{weightsreparamBN}
\tilde{W}_{1} =  \alpha W_{1}\quad {\rm and}\quad \tilde{W}_{2} =  \beta W_{2}, 
\end{align}
where $\mathbf{\alpha} = \Diag(\alpha_{1},\alpha_{2},\alpha_{3},\alpha_{4},\alpha_{5},\alpha_{6},\alpha_{7},\alpha_{8}) $ and $ \mathbf{\beta} = \Diag(\beta_{1},\beta_{2},\beta_{3},\beta_{4},\beta_{5},\beta_{6},\beta_{7},\beta_{8})$
and the elements of $\alpha,\beta$ can be any real number. This loss is invariant to this reparameterization of the weights as can be seen by following a similar derivation shown for Arch1. \changeBM{It should be noted that the additional parameters used in Arch2, proposed in \citep{BN}, are left unchanged.}

Unfortunately, the Euclidean gradient of the weights used in standard SGD weight update is not invariant to these reparameterizations of the weights such as those possible in Arch1 and Arch2. This can be seen \changeBM{in the simple example of a function $f : \mathbb{R} \rightarrow \mathbb{R}: x \mapsto f(x)$ that is invariant under the transformation $\tilde{x} = \alpha x$ for all non-zero scalar $\alpha$, i.e., $f(\tilde{x}) = f(x)$. Equivalently,}
\begin{equation}\label{EuclideanGradient}
\begin{array}{ll}
\displaystyle \frac{\partial f(\tilde{x})}{\partial \tilde{x}} =  \frac{\partial f(\tilde{x})}{\partial x} \frac{\partial x}{\partial \tilde{x}}  =  \frac{\partial f(x)}{\partial x} \frac{1}{\alpha},
\end{array}
\end{equation}
\changeBM{where $ {\partial f(x)}/{\partial x}$ is the Euclidean gradient of $f$ at $x$. As is clear in (\ref{EuclideanGradient}), the Euclidean gradient is not invariant to reparameterizations, i.e., it scales \emph{inversely} to the scaling of the variable $x$. Consequently, the optimization trajectory can vary significantly based on the chosen parameterization. On the other hand, a scale-invariant gradient scales \emph{proportionally} to that of the scaling of the variable.} This issue can be resolved either by defining a suitable non-Euclidean gradient which is invariant to reparameterizations or by placing appropriate constraints on the filter weights as we show in the following section \citep[Chapter~3]{absil08a} .




\section{Resolving symmetry issues using manifold optimization}
\label{Optimization}

\changeBM{We propose two ways of resolving the symmetries that arise in deep architectures. First, we follow the approach of \citep{Amari, edelman98a, absil08a} to equip the search space with a new non-Euclidean metric to resolve the symmetries present. Second, we break the symmetries by forcing the filter weights to be on the unit-norm manifold. In both these case, our updates are simple to implement in a stochastic gradient descent setting on manifolds \citep{bonnabel13a}.}  \changeVB{The proposed updates are shown for a two layer deep network. However, the updates can be readily extended to deeper architectures.}

Consider a weight vector $w \in \mathbb{R}^{n}$, the Euclidean squared length of a small incremental vector $dw$ connecting $w$ and $w+dw$ is given by
\begin{align}
\label{EuclideanMetric}
\| dw \|^{2} = \sum_{i=1}^{n} ( dw_{i} )^{2}.
\end{align}
For a non-Euclidean coordinate system, however, the notion of squared distance is given by \changeBM{the Riemannian metric}
\begin{align}
\label{RiemannianMetric}
\| dw \|_G^{2} = dw^{T} G dw, 
\end{align}
where the matrix $G$ is a positive definite matrix. \changeBM{If $G $ is the identity matrix,} then the coordinate system in (\ref{RiemannianMetric}) is Euclidean. The steepest descent direction for a loss function $\ell(w)$ under the metric (\ref{RiemannianMetric}) is given by $\tilde{\nabla} \ell(w) = G^{-1}\nabla \ell(w)$, \changeBM{where $\nabla \ell(w)$ is the Euclidean gradient and $\tilde{\nabla} \ell(w)$ is the Riemannian gradient under the metric (\ref{RiemannianMetric}).} Consequently, the first order weight update is of the form
\begin{equation}\label{SM}
\begin{array}{lll}
\tilde{\nabla} \ell(w) &=& G^{-1}\nabla \ell(w) \\
w^{t+1} &=& w^{t} - \lambda \tilde{\nabla} \ell(w),
\end{array}
\end{equation}
\changeBM{where $w^t$ is the current weight, $\nabla \ell(w)$ is the Euclidean gradient, $w^{t+1}$ is the updated weight, and $\lambda$ is the learning rate. Therefore, to resolve the symmetries for $(W_1, W_2, \theta)$ discussed in Section \ref{Architecture}, we propose the \emph{novel} Riemannian metric}
\begin{equation}
\label{MetricDef}
\begin{array}{lll}
\| d{W}_1\|^2_{G_{W_{1}}} & =& \Trace( d{W}^T_1 G_{W_{1}}   d{W}_1) \\
\| d{W}_2\|^2_{G_{W_{2}}} & =& \Trace( d{W}^T_2 G_{W_{2}}   d{W}_2) \\

\changeVB{ \| d{\theta}\|^2_{\theta}} &= & \changeVB{ \Trace( d{\theta}^{T} d{\theta} G_{\theta}   )  }, \\
\\
{\rm where} & & \\
G_{W_{1}} &=&  (\Diag ( \diag( W_{1}W_{1}^{T} )))^{-1}, \\
G_{W_{2}} &= &( \Diag ( \diag( W_{2}W_{2}^{T})))^{-1}, {\rm and}\\
\changeVB{G_{\theta} } &=& \changeVB{ (\Diag( \diag(\theta^{T} \theta)))^{-1} }.
\end{array}
\end{equation}
Here $\Trace(\cdot)$ takes the trace of a square matrix, $\Diag(\cdot)$ is an operator which creates a diagonal matrix with its argument placed along the diagonal, and $\diag(\cdot)$ is an operator which extracts the diagonal elements of the argument matrix. \changeBM{It should be noted that $G$ is defined separately for the weights $W_1$, $W_2$, and $\theta$ in (\ref{RiemannianMetric}), which is invariant to the reparameterizations shown in (\ref{weightsreparam}) and (\ref{thetareparam}) for Arch1 and (\ref{weightsreparamBN}) for Arch2. It should be noted that $G_{\theta}$ in (\ref{MetricDef}) acts from the right side in the case of $\theta$. The motivation behind the metric choice in (\ref{MetricDef}) comes from the classical notion of \emph{right} and \emph{left} invariances in differential geometry, but now restricted to diagonal elements. To show the invariance of the proposed metric, we consider the invariance for $\theta$ in Arch1 as an example.}
\changeVB{
\[
\begin{array}{lll}
\| d\tilde{\theta} \|_{G_{\tilde{\theta}}}^{2}  =  \Trace(d\tilde{\theta} G_{\tilde{\theta}} d\tilde{\theta}^{T})\\
 =   \Trace(( d\theta \frac{1}{\alpha_{0}} \beta_{s}^{-1} )( \Diag( \diag(\frac{1}{\alpha_{0}} \beta_{s}^{-T} \theta^{T} \theta \frac{1}{\alpha_{0}} \beta_{s}^{-1} )  ))^{-1} \\
 \quad ( \frac{1}{\alpha_{0}} \beta_{s}^{-T} d\theta^{T} )) \\
 =   \Trace(d\theta \frac{1}{\alpha_{0}} \beta_{s}^{-1} \beta_{s} \alpha_{0} (\Diag( \diag( \theta^{T} \theta ) ))^{-1} \alpha_{0} \beta_{s}^{T} \beta_{s}^{-T} \frac{1}{\alpha_{0}} d\theta^{T}) \\
=   \Trace(d\theta (\Diag( \diag( \theta^{T} \theta )) )^{-1} d\theta^{T}) \\
=   \| d{\theta} \|_{G_\theta}^{2},
\end{array}
\]
}
and therefore, the squared length (\ref{RiemannianMetric}) is left unchanged (i.e., the metric is invariant) under the considered reparameterization of $\theta$ in (\ref{thetareparam}). Similar derivations show that the metric in (\ref{MetricDef}) is invariant to reparameterizations in $W_{1}$ and $W_{2}$. We term the \changeBM{proposed} metric in (\ref{MetricDef}), collectively as the \textit{scaled metric} (SM). The scaled metric SM is equally applicable as an invariant metric for Arch2 which possesses symmetries shown in (\ref{weightsreparamBN}).

\begin{table*}[t]%
\caption{Symmetry-invariant updates}
\label{ProposedUpdates}
\center \small
\begin{tabular}{c|c}
 Scaled metric (SM) & Unit norm (UN) \\
\hline
 & \\
 $\begin{array}[t]{lll}
W_1^{t+1} = W_1^t - \lambda  G_{W_{1}^t}^{-1} {\nabla}_{W_1} L (W_1^t, W_2^t, \theta^t)
\end{array} $ &  $\begin{array}[t]{lll}
\tilde{{\nabla}}_{W_1} L (W_1^t, W_2^t, \theta^t) = \Pi_{W_1^t}({{\nabla}}_{W_1} L (W_1^t, W_2^t, \theta^t))\\
W_1^{t+1} = {\Orth}({W_1^t} - \lambda \tilde{{\nabla}}_{W_1} L (W_1^t, W_2^t, \theta^t))
\end{array} $ \\
 & \\
 $\begin{array}[t]{lll}
W_2^{t+1} = W_2^t - \lambda  G_{W_{2}^t}^{-1} {\nabla}_{W_2} L (W_1^t, W_2^t, \theta^t)
\end{array} $ & $\begin{array}[t]{lll}
\tilde{{\nabla}}_{W_2} L (W_1^t, W_2^t, \theta^t) = \Pi_{W_2^t}({{\nabla}}_{W_2} L (W_1^t, W_2^t, \theta^t))\\
W_2^{t+1} = {\Orth}({W_2^t} - \lambda \tilde{{\nabla}}_{W_1} L (W_1^t, W_2^t, \theta^t))
\end{array} $  \\
& \\
 $\begin{array}[t]{lll}
\theta^{t+1} = \theta^t - \lambda  {\nabla}_{\theta} L (W_1^t, W_2^t, \theta^t) G_{{\theta}^t}^{-1} 
\end{array} $ & $\begin{array}[t]{lll}
\theta^{t+1} = \theta^t - \lambda  {\nabla}_{\theta} L (W_1^t, W_2^t, \theta^t)
\end{array} $ \\
 & \\
\hline
\end{tabular}%
\end{table*}

Another way to \changeBM{resolve} the symmetries that exist in Arch1 and Arch2 is to constrain the weight vectors (filters) in $W_{1}$ and $W_{2}$ to lie on the \emph{oblique manifold} \citep{absil08a, ManOpt}, i.e., each filter in the fully connected layers is constrained to have \changeBM{unit} Euclidean norm. \changeBM{Equivalently, we impose the constraints $\diag (W_1 W_1^T) = 1$ and $\diag (W_2 W_2^T) = 1$, where $\diag(\cdot)$ is an operator which extracts the diagonal elements of the argument matrix.} 

Consider a weight vector $w \in \mathbb{R}^{n}$ with the constraint $w^Tw = 1$. (For example, $w^T$ is a row of $W_1$.) The steepest descent direction for a loss $\ell(w)$ \changeBM{with $w$ on the unit-norm manifold} is computed $\tilde{\nabla} \ell(w) = \nabla \ell(w) - (w^{T}\nabla \ell(w)) w$, \changeBM{where $\nabla \ell(w)$ is the Euclidean gradient and $\tilde{\nabla} \ell(w)$ is the Riemannian gradient on the unit-norm manifold \citep[Chapter~3]{absil08a}.} Effectively, the normal component of the Euclidean gradient, i.e., $(w^{T}\nabla \ell(w)) w$, is subtracted to result in the tangential (to the unit-norm manifold) component. \changeBM{Following the tangential direction takes the update out of the manifold, which is then pulled back to the manifold with a \emph{retraction} operation \citep[Example~4.1.1]{absil08a}.} \changeBM{Finally}, an update of the weight $w$ on the unit-norm manifold is of the form
\begin{equation}\label{UN}
\begin{array}{lll}
\tilde{\nabla} \ell(w) &= & \nabla \ell(w) - (w^{T}\nabla \ell(w)) w \\
\tilde{w}^{t+1}& = & w^{t} - \lambda \tilde{\nabla} \ell(w) \\ 
w^{t+1} &= & \tilde{w}^{t+1} / \| \tilde{w}^{t+1} \|,
\end{array}
\end{equation}
where $w^t$ is the current weight, $\nabla \ell(w)$ is the Euclidean gradient, $w^{t+1}$ is the updated weight, and $\lambda$ is the learning rate. \changeBM{It should be noted when $W_1$ and $W_2$ are constrained, the $\theta$ variable is reparameterization free.}

Both the proposed weight updates, \changeBM{(\ref{SM}) that is based on the scaled metric (SM) and (\ref{UN}) that is based on the unit-norm (UN) constraint}, can be used in a stochastic gradient descent (SGD) setting which we use in our experiments described in the following section. It should be emphasized that the proposed updates are numerically efficient to implement. \changeVB{The Euclidean gradients are computed efficiently using gradient back-propagation \citep{rumelhart1988learning}.}

\changeBM{The proposed symmetry-invariant updates for a loss function $L(W_1, W_2, \theta)$ in Arch1 and Arch2 type networks are shown in Table \ref{ProposedUpdates}. Here $(W_1^t, W_2^t, \theta ^t)$ is the current weight, $(W_1^{t+1} , W_2^{t+1} , \theta^{t+1} )$ is the updated weight, $\lambda$ is the learning rate, and ${\nabla}_{W_1} L (W_1^t, W_2^t, \theta^t)$,  ${\nabla}_{W_2} L (W_1^t, W_2^t, \theta^t)$, and ${\nabla}_{\theta} L (W_1^t, W_2^t, \theta^t)$ are the partial derivatives of the loss $L$ with respect to $W_1$, $W_2$, and $\theta$, respectively at $(W_1^t, W_2^t, \theta ^t)$. The matrices $G_{W_{1}^t}$, $G_{W_{2}^t}$, and $G_{\theta ^t}$ are defined in (\ref{MetricDef}) at $(W_1^t, W_2^t, \theta ^t)$. The operator $\Orth(\cdot)$ normalizes the rows of the input argument. $\Pi_W(\cdot)$ is the linear projection operation that projects an arbitrary matrix onto the tangent space of the oblique manifold at an element $W$. Specifically, it is defined as $\Pi_W(Z) = Z - \Diag(\diag((ZW^T))W$ \citep{ManOpt}, where $\Diag(\cdot)$ is an operator which creates a diagonal matrix with its argument placed along the diagonal and $\diag(\cdot)$ is an operator which extracts the diagonal elements of the argument matrix. The additional parameters used in Arch2 are updated as proposed in \citep{BN}. It should be noted that $G_{\theta}^{-1}$ acts from the right side in the update of $\theta$.}

The convergence analysis of SGD on manifolds follows the developments in \citep{Bottou, bonnabel13a}.

%
%
%
%


\section{Experimental setup}
\label{Experiments}

We train both two and four layer deep Arch1 and Arch2 networks to perform digit classification on the MNIST dataset. This dataset has $60000$ training images and $10000$ testing images. For both \changeVB{these architectures} we use 64 features per layer. The digit images are rasterized into a $784$ dimensional vector as input to the network(s). No input pre-processing \changeBM{is} performed. The weights in each layer are drawn from a standard Gaussian and each filter is unit-normalized. The soft-max class vectors are also drawn from a standard Gaussian and each class vector is unit-normalized.  

We use stochastic gradient descent based optimization as this is the most widely used technique for training deep networks. The three different weight updates \changeBM{that} we compare within this first-order framework are scaled metric (SM), unit-norm (UN), and balanced SGD (B-SGD). \changeVB{B-SGD uses the Euclidean updates, but wherein the starting values of filters and class vectors are unit-normalized.} B-SGD is also studied \changeBM{as a benchmark algorithm} in \citep{PathSGD}. We choose a mini-batch size \changeBM{of} $100$ samples.

We choose the base learning rate from the set $10^{-p}$ \changeBM{for $p \in \{2, 3, 4, 5\}$} for each training run of the experimental network. \changeVB{To select the optimal learning rate from this set, we create a validation set of $500$ images from the training set for testing. We then train the network with each learning rate using a randomly chosen set of $1000$ images from the training set for $50$ epochs. At the start of each epoch, the training set is randomly permuted and mini-batches are sampled in a sequence ensuring each training sample is used only once within an epoch. We record the error on the validation set measured as the error per validation sample for each candidate base learning rate. Then the candidate rate which corresponds to the lowest validation error is selected and used for training the network on the full training set. We repeat this process of learning rate selection and training of the network with the full training set $10$ times for each of the three weight update strategies. For each of these runs, we measure the mean and variance of the test error. \changeBM{We ignore a small proportion of runs where the validation error diverged.}}

\begin{table*}[t]%
\caption{Comparisons on the MNIST dataset}
\label{TestErr}
\begin{center}
\scriptsize{
\begin{tabular}{|c|c|c|c|c|c|c|}
\hline
& \multicolumn{3}{|c|}{Arch1} & \multicolumn{3}{|c|}{Arch2}  \\
\hdashline
Protocol & B-SGD & SM & UN & B-SGD & SM & UN \\
\hline
\multicolumn{7}{|c|}{2 layer deep network} \\
\hdashline
Exp. decay & 0.0263 $\pm$ 0.0079 & 0.0283 $\pm$ 0.0062 & \textbf{0.0220} $\pm$ 0.0057 & \textbf{0.0216} $\pm$ 0.0038 & 0.0228 $\pm$ 0.0051 & 0.0230 $\pm$ 0.0036  \\
\hdashline
Bold driver & 0.0240 $\pm$ 0.0026 & 0.0271 $\pm$ 0.0076 & \textbf{0.0228} $\pm$ 0.0038 & 0.0206 $\pm$ 0.0024 & \textbf{0.0186} $\pm$ 0.0024 & 0.0199 $\pm$ 0.0046 \\
\hline
\multicolumn{7}{|c|}{4 layer deep network}\\
\hdashline
Exp. decay & 0.0277 $\pm$ 0.0045 & 0.0256 $\pm$ 0.0038 & \textbf{0.0215} $\pm$ 0.0049 & 0.0218 $\pm$ 0.0028 & \textbf{0.0204} $\pm$ 0.0050 & 0.0224 $\pm$ 0.0065 \\
\hdashline
Bold driver & 0.0253 $\pm$ 0.0060 & 0.0264 $\pm$ 0.0056 &\textbf{ 0.0244} $\pm$ 0.0101 & 0.0204 $\pm$ 0.0027 & 0.0188 $\pm$ 0.0033 & \textbf{0.0179} $\pm$ 0.0025 \\
\hline
\end{tabular}}%
\end{center}
\end{table*}

For training each network, we use the two well known protocols for annealing or decaying the learning rate; the bold-driver \changeVB{(annealing)} protocol \citep{HintonBoldDriver} and the exponential decay protocol. For the exponential decay protocol, we choose a decay factor of $0.95$ after each epoch. In all, for each network, we use two protocols, three different weight update strategies, and 10 training runs for each combination thus totaling sixty training runs.

For each training run on the full dataset, we choose $50000$ randomly chosen samples as the training set and the remaining $10000$ samples for validation. We train for a minimum of $25$ epochs and a maximum of $60$ epochs. \changeVB{When the bold driver protocol is used, we terminate the training if, (i) the training error is less than $10^{-5}$, (ii) the validation error increases with respect to the one measured $5$ epochs earlier, (iii) successive validation error measurements differ less than $10^{-5}$. }


\section{Results and analysis}\label{Results}


The mean and standard deviation of the test error for various training combinations are tabulated in Table \ref{TestErr}. From these quantitative figures we can observe that Arch2 performs significantly better than Arch1. This emphasizes the ability of batch normalization to improve performance \citep{BN}. Arch1 results are characterized by high mean and large standard deviation values for all three weight updates. However, there is no clear indication of the superiority of one protocol over the other. While the bold driver protocol improves the performance of B-SGD, it worsens the performance of SM and UN in the four layer deep network. The exponential decay protocol produces the best result in combination with the UN weight updates and has better performance for SM and UN for the four layer deep network. \changeVB{The \changeBM{good} performance of UN can be explained by the fact that it allows for better gradient back-propagation \citep{rumelhart1988learning} as the norm of the filters are constrained to be unit norm.} 

The Arch2 network which includes batch normalization \citep{BN} helps improve gradient back-propagation by constraining the scales of the input \changeVB{feature maps} to each layer. The beneficial result of adding this layer is clearly visible from our results where we see both SM and UN perform much better than B-SGD with the bold driver protocol. Both the mean and \changeVB{standard deviation} of the test error are the lowest for SM and UN when both the two and four layer deep networks are considered. The bold driver protocol performs better than the exponential decay protocol for all the considered weight updates. The annealing protocol seems more suitable when sufficient regularization is in place as in Arch2 with SM and UN.

We show the mean trajectories of the test error over the training epochs for the four layer deep Arch1 and Arch2 in Figure \ref{Trajectory}. For Arch1, we show the results using the exponential decay protocol and for Arch2 we show results using the bold driver protocol. \changeVB{In practice,} when using bold driver, the training is terminated based on the stopping criterion at most after $30$ epochs.

\begin{table*}[th]%
\begin{tabular}{cc}
\includegraphics[width=0.48\textwidth]{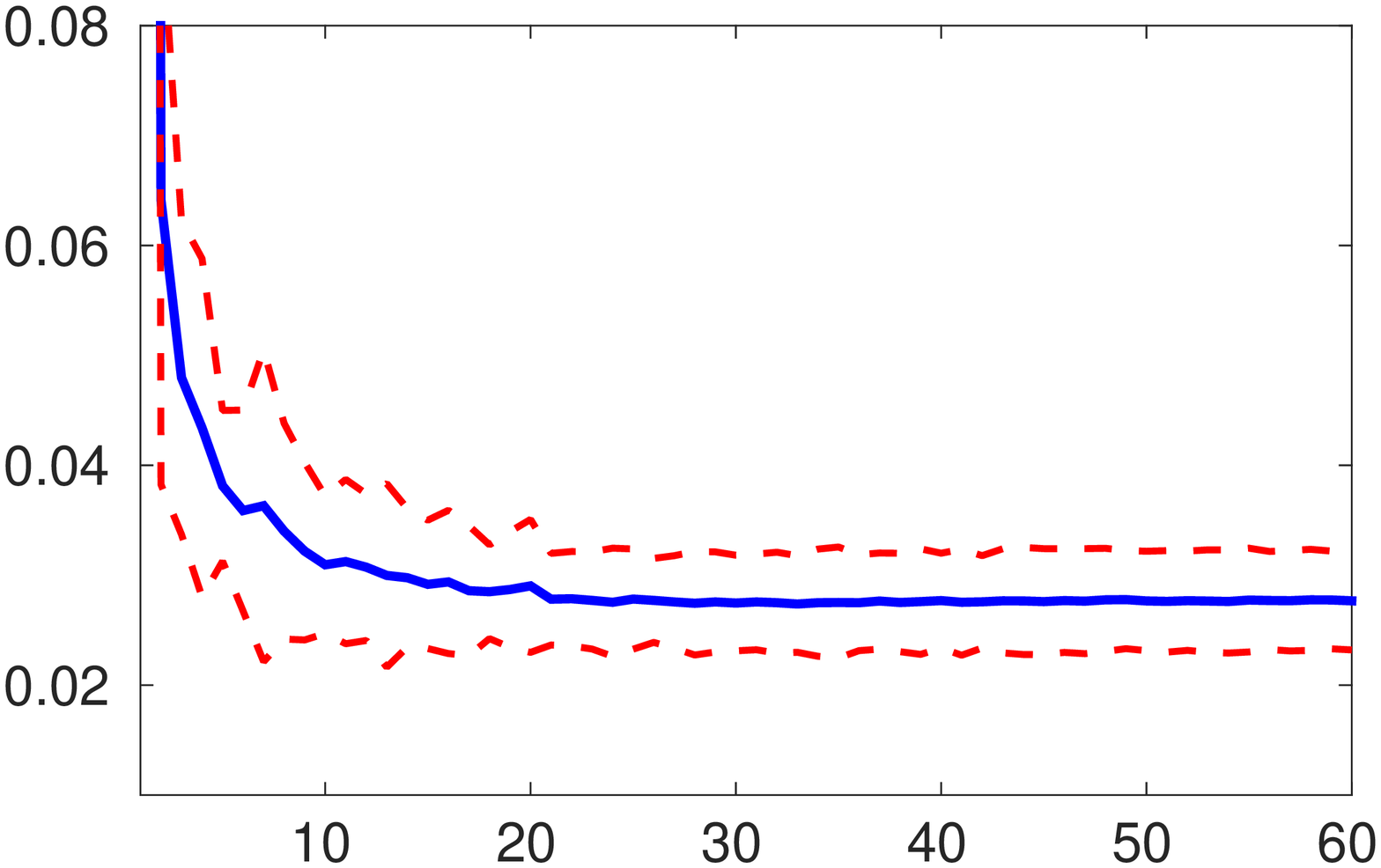}
&
\includegraphics[width=0.48\textwidth]{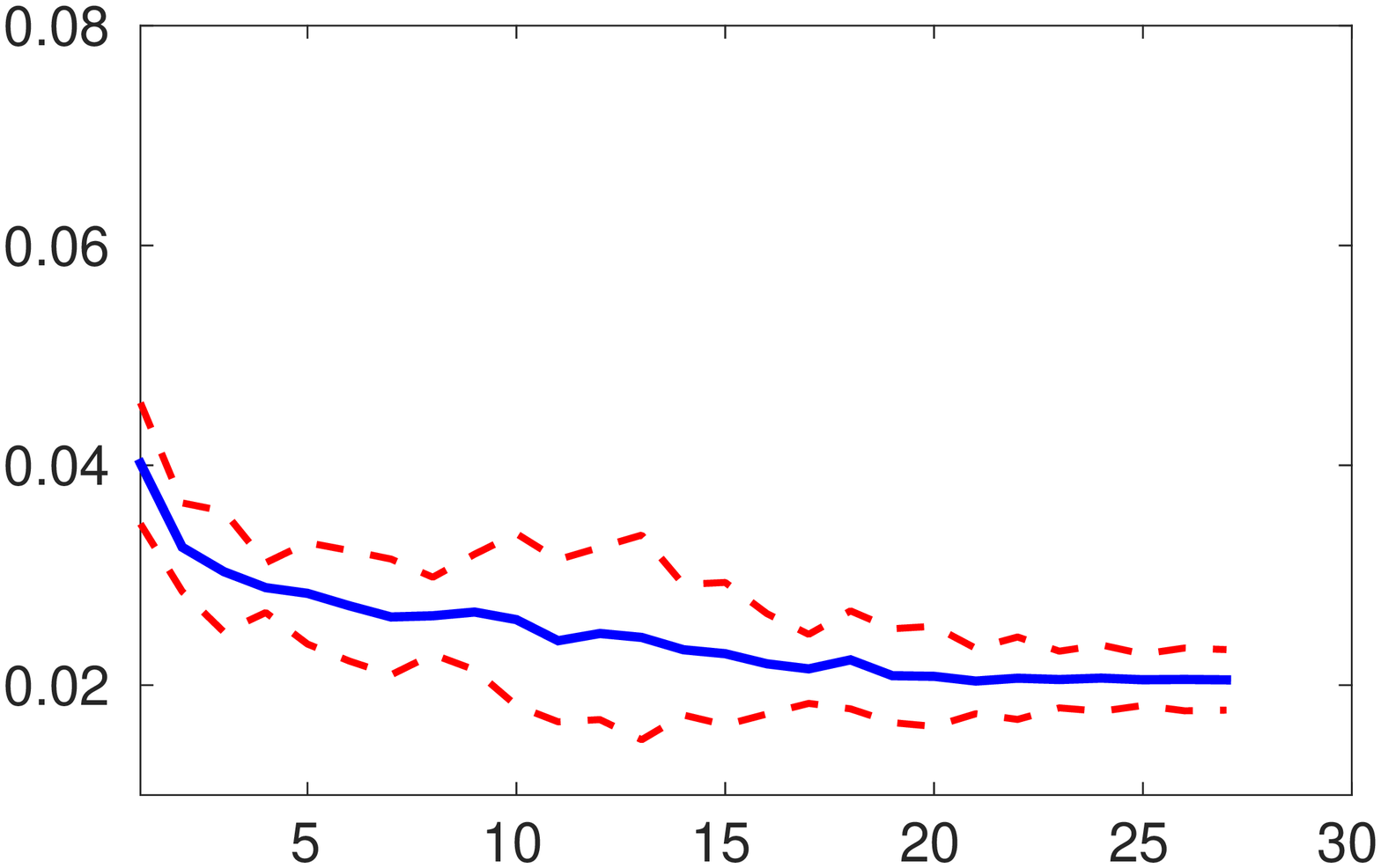}
\\
\footnotesize{Arch1: B-SGD Exp. decay.} & \footnotesize{Arch2: B-SGD Bold driver.}
\\
\includegraphics[width=0.48\textwidth]{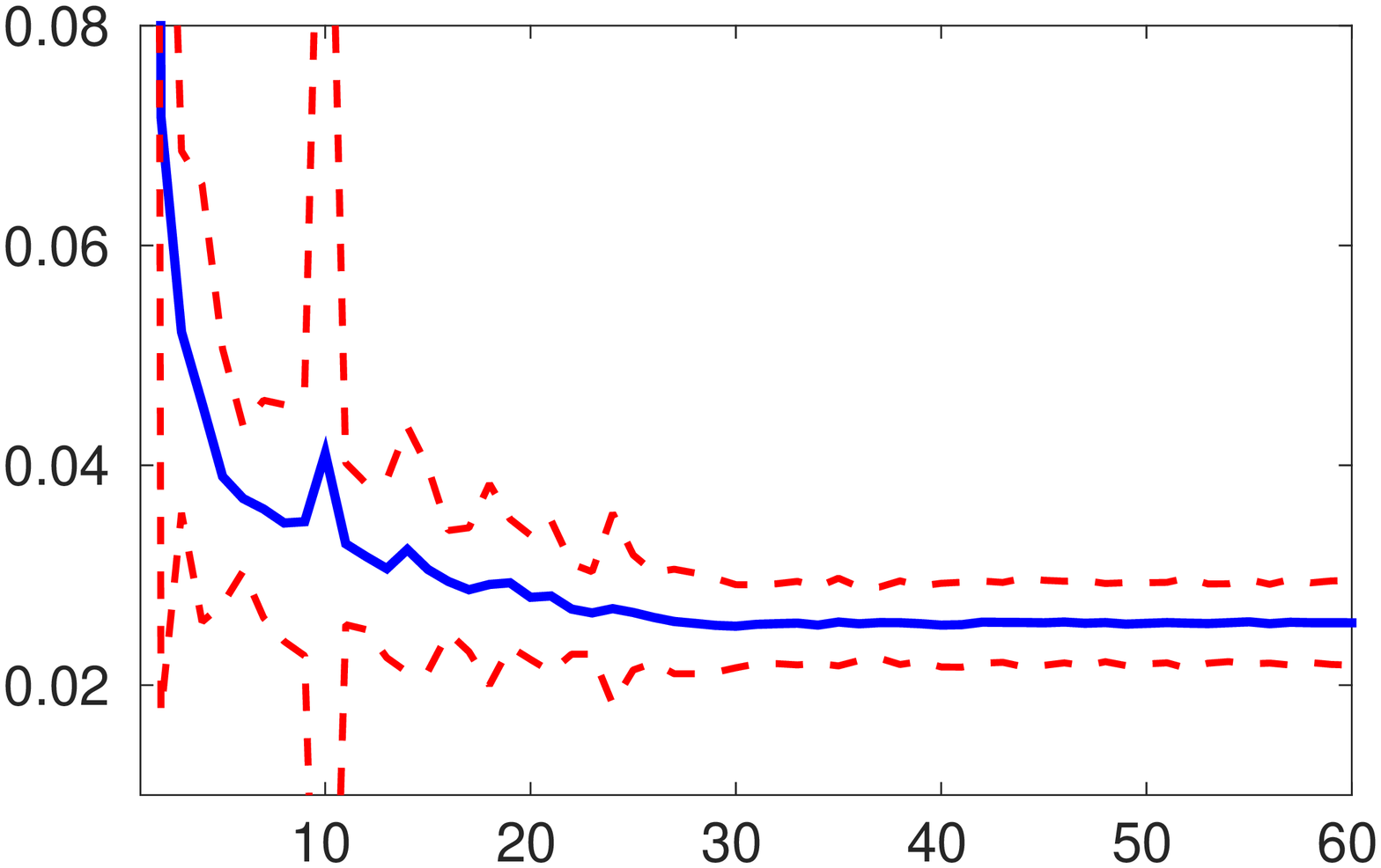}
&
\includegraphics[width=0.48\textwidth]{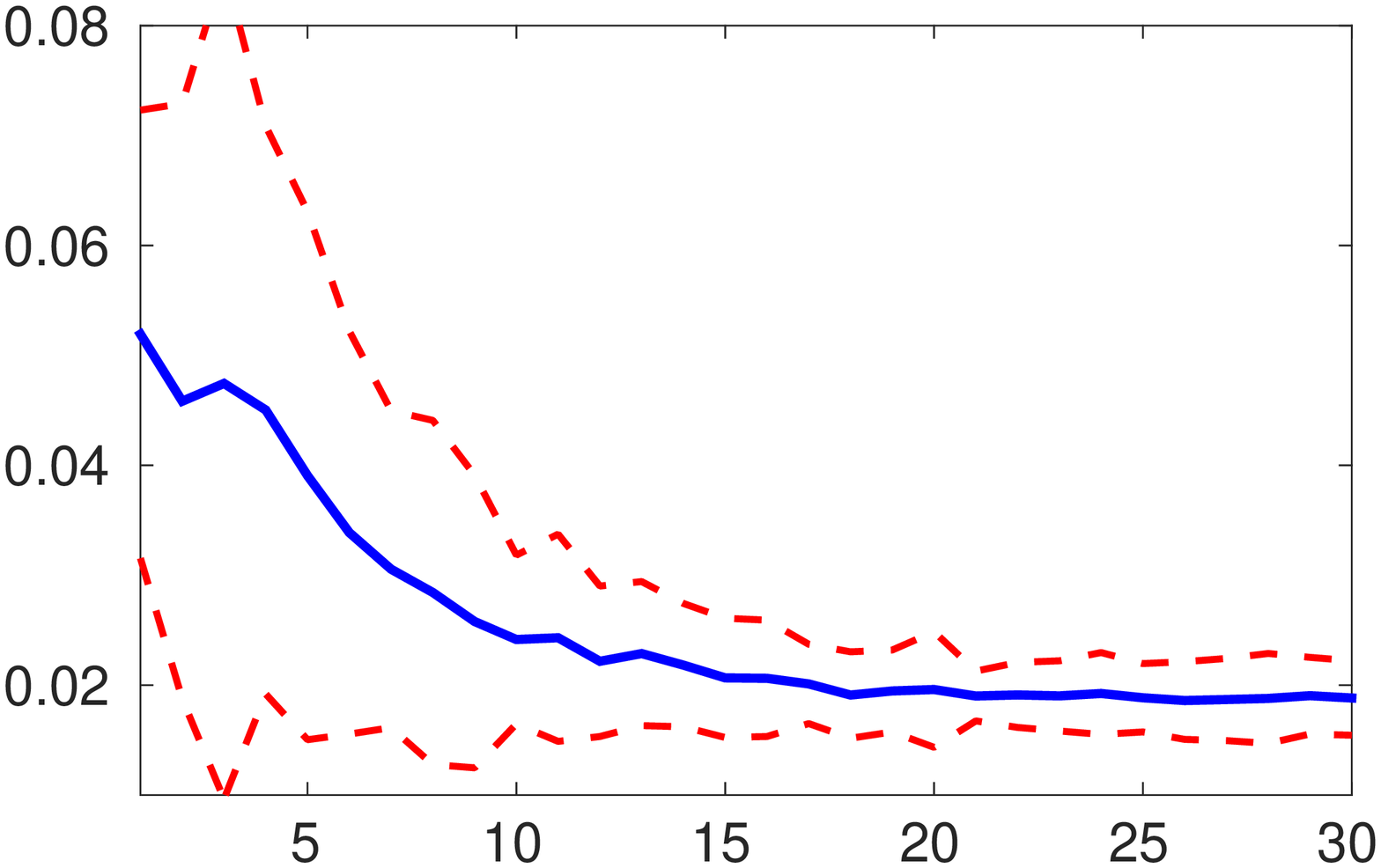}\\
\footnotesize{Arch1: SM Exp. decay.} & \footnotesize{Arch2: SM Bold driver.}
\\
\includegraphics[width=0.48\textwidth]{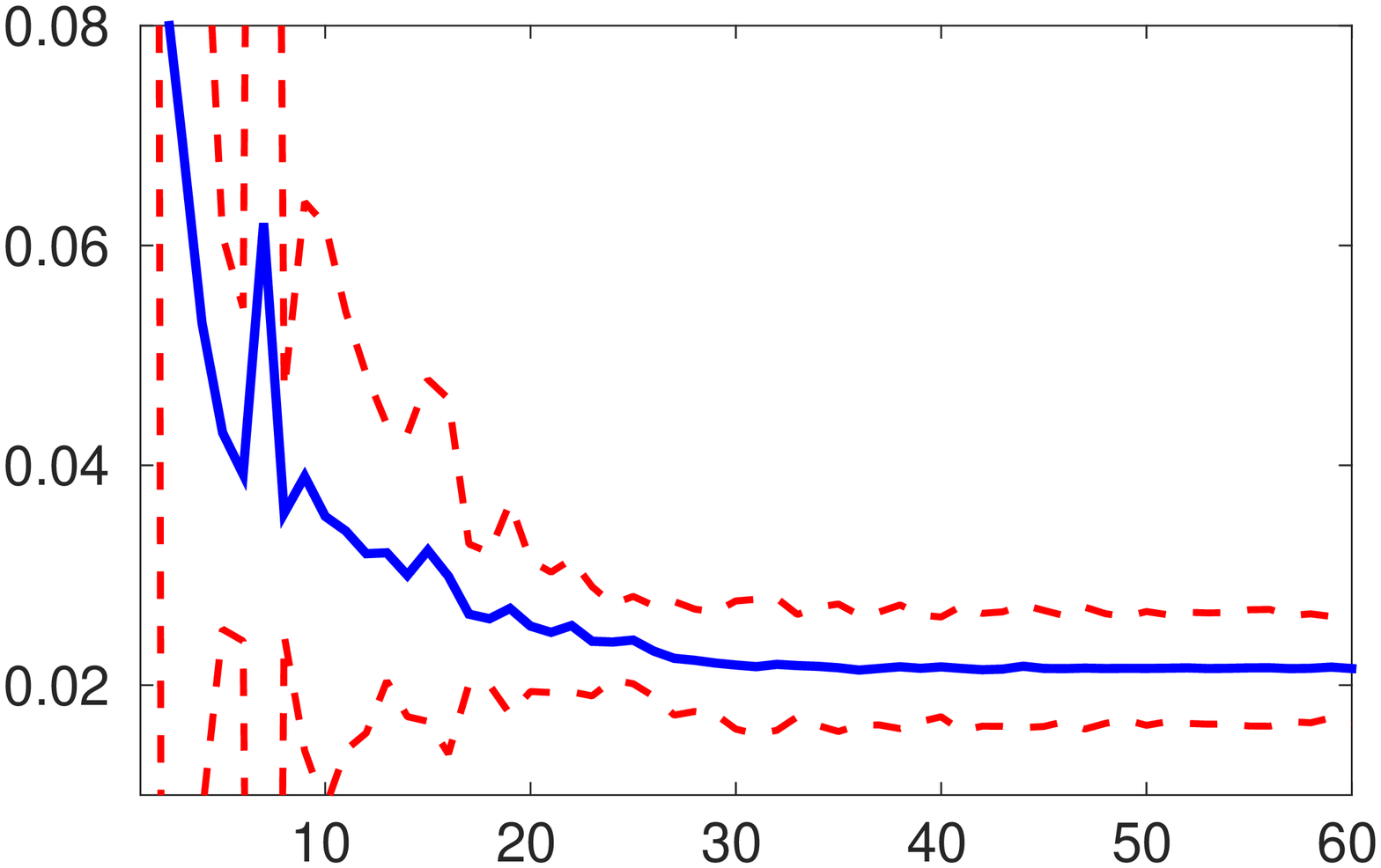}
&
\includegraphics[width=0.48\textwidth]{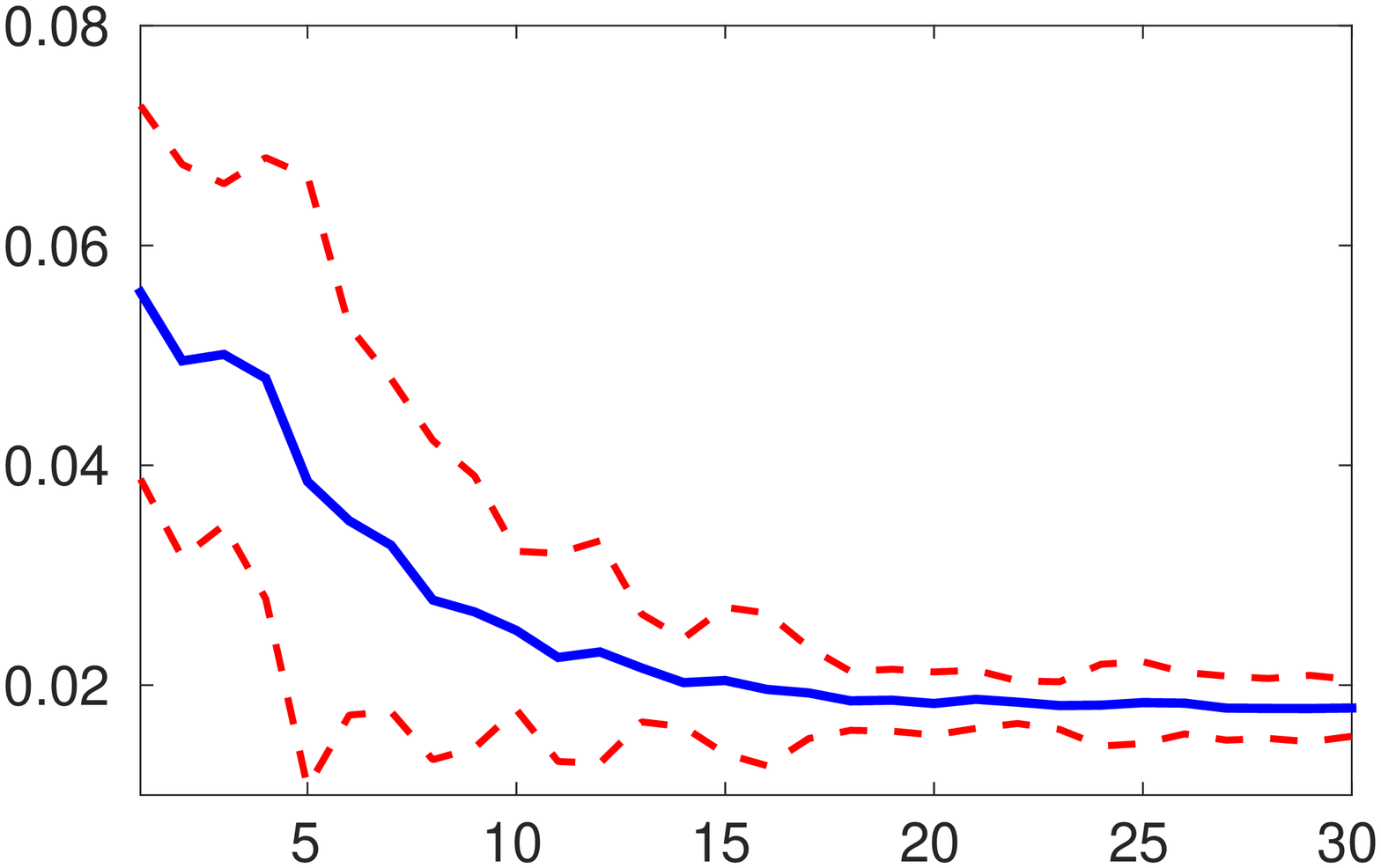}\\
\footnotesize{Arch1: UN Exp. decay.} & \footnotesize{Arch2: UN Bold driver.}
\end{tabular}
\caption{Trajectories for four layer deep Arch1 and Arch2 networks which result in the lowest mean test error. The one standard deviation band is shown along with the mean trajectory dotted lines.}
\label{Trajectory}
\end{table*}


\changeBM{The best results \changeBM{are} obtained using Arch2 with SM and UN weight updates and with the bold driver learning rate annealing protocol.} SM and UN absorb the symmetries present in Arch2 and help improve performance over what is achieved by \changeBM{standard} batch normalization alone. The use of SM and UN can also be seen as \changeBM{a} way to regularize the weights of the network during training without introducing any hyper-parameters, \changeBM{e.g., a weight decay term}.

The quantitative results show that SM for a two layer deep network with the bold driver protocol performs better than using the B-SGD update for training a four layer deep network with exponential decay of the learning rate. It is also noteworthy to observe that the performance difference between the two and four layer deep Arch2 network is not very large. This raises the question for future research as to whether some of these networks necessarily have to be that deep \citep{Caruana} or it can be made shallower (and efficient) by better optimization.


\section{Application to image segmentation}
\label{DeepConvNets}

\begin{figure*}[t]
\centering
\includegraphics[width=1\textwidth]{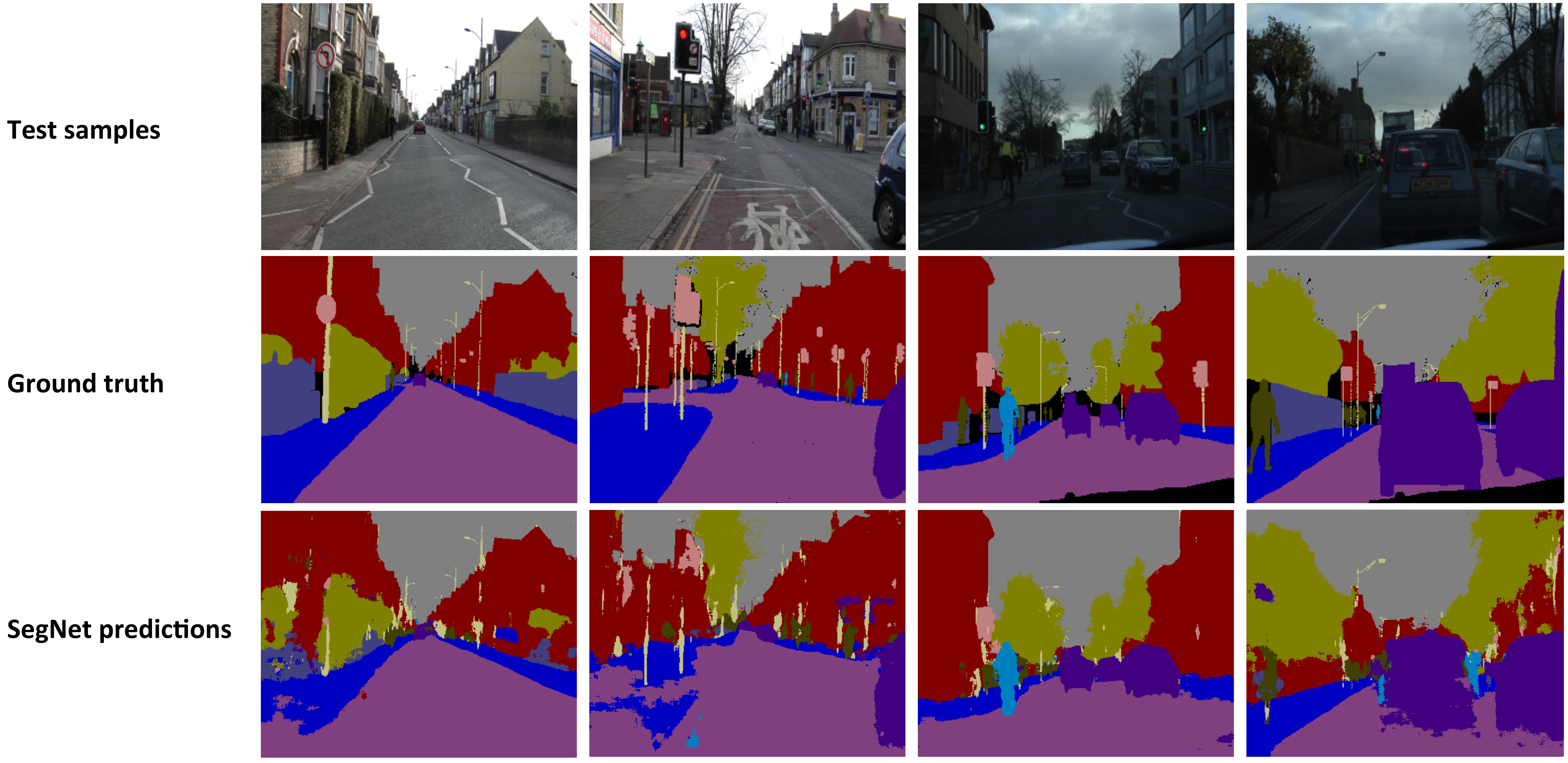}
\caption{Using SGD with the proposed UN weight update, shown in Table \ref{ProposedUpdates}, for training SegNet \citep{SegNetarXiv}. The quality of the predictions as compared to the ground truth indicates a successful training of the network.}
\label{CamVidQualy}
\end{figure*}

We \changeBM{apply SGD with the proposed UN weight updates in Table \ref{ProposedUpdates} for} training SegNet, a deep convolutional network proposed for road scene image segmentation into multiple classes \citep{SegNetarXiv}. This network, although convolutional, possesses the same symmetries as those analyzed for Arch2 in (\ref{weightsreparamBN}). The network \changeBM{is} trained for 100 epochs on the CamVid \citep{GabeDataset} training set of 367 images. The predictions of the trained SegNet on some sample test images from the dataset can be seen in Figure \ref{CamVidQualy}. These qualitative results indicate the usefulness of our analysis and \changeBM{symmetry-invariant} weight updates for larger networks that arise in practice.


\section{Conclusion}
\label{Conclusion}
\changeVB{We have highlighted the symmetries that exist in the weight space of currently popular deep neural network architectures}. We have shown that these symmetries can be \changeVB{handled well} in stochastic gradient descent optimization framework either by designing an appropriate non-Euclidean metric or by \changeBM{imposing} a unit-norm constraint on the filter weights. Both of these strategies take into account the manifold structure on which the weights of the network reside \changeBM{and lead to symmetry-invariant weight updates}. The empirical results show the test performance can be improved using our proposed symmetry-invariant weight updates even on modern architectures. \changeBM{As a future research direction,} we would exploit these techniques for deep convolutional neural networks \changeBM{used in} practical applications.

\subsubsection*{Acknowledgments}
Vijay Badrinarayanan and Roberto Cipolla were supported by a sponsorship from Toyota Motor Europe, Belgium. Bamdev Mishra was supported as an FNRS research fellow (Belgian Fund for Scientific Research). The scientific responsibility rests with its authors.

\bibliography{RefBase}
\bibliographystyle{iclr2016_conference}

\end{document}